\documentclass[letterpaper, 10 pt, conference]{ieeeconf}  % Comment this line out if you need a4paper

\usepackage{stix}
\usepackage{pifont}% http://ctan.org/pkg/pifont
\usepackage{pbox}

\IEEEoverridecommandlockouts                              % This command is only needed if 
\title{\LARGE \textbf{Toward Teach and Repeat Across Seasonal Deep Snow Accumulation}}

\author{Matěj Boxan*$^{1}$, Alexander Krawciw*$^{2}$, Timothy D. Barfoot$^{2}$, and François Pomerleau$^{1}$% <-this % stops a space
\thanks{$^{1}$Northern Robotics Laboratory, Université Laval, Quebec~City, Canada
		{\texttt{\footnotesize\{matej.boxan, francois.pomerleau\}@norlab.ulaval.ca}}
}
\thanks{$^{2}$University of Toronto Robotics Institute, Toronto, Canada
    {\texttt{\footnotesize\{alec.krawciw, tim.barfoot\}@robotics.utias.utoronto.ca}}
}%
\thanks{* denotes equal contribution.}
\thanks{This research was supported by the Natural Sciences and Engineering Research Council of Canada (NSERC) and the Fonds de recherche du Québec (FRQNT) through the grant 2023-NOVA-326877 HUNTER (Highlight the Unexpected with Navigation Through Extreme Regions). 
The weather data were acquired from the \emph{Adaptable Earth Observation System} project funded by the Canada Foundation for Innovation, the Government of Quebec, McGill, and UQAM.
A. Krawciw is supported by a Vanier Canada Graduate Scholarship}% <-this % stops a space
}

\usepackage[l2tabu,orthodox]{nag}

\usepackage[
    backend=bibtex8,
    style=ieee,
    sorting=none,
    natbib=true,
    doi=false,
    isbn=false,
    url=false,
    eprint=false,
    maxcitenames=1,
    mincitenames=1
]{biblatex}

\usepackage[pdftex,colorlinks]{hyperref}
\usepackage[printonlyused]{acronym}

\usepackage{siunitx}
\usepackage[all]{nowidow}

\usepackage[dvipsnames]{xcolor}

\usepackage{lipsum}

\usepackage{amssymb,amsfonts,amsmath,amscd}

\usepackage[french,english,noabbrev,nameinlink]{cleveref}

\usepackage{bm}

\newcommand{\bbm}{\begin{bmatrix}}
\newcommand{\ebm}{\end{bmatrix}}
\usepackage{xspace} %smart handling of space in commands

\usepackage[pdftex]{graphicx}

\usepackage{epstopdf}

\usepackage{import}

\usepackage{overpic}

\graphicspath{{./latexGoodPractices/}}

\usepackage{booktabs}

\usepackage{tabularx}
\usepackage{multirow, multicol}

\acrodef{UGV}{Uncrewed Ground Vehicle}
\acrodef{IMU}{Inertial Measurement Unit}
\acrodef{MEMS}{Micro-Electromechanical Systems}
\acrodef{GNSS}{Global Navigation Satellite System}
\acrodef{PTP}{IEEE1588 Precision Time Protocol}
\acrodef{SLAM}{Simultaneous Localization and Mapping}
\acrodef{DOF}{Degrees of Freedom}
\acrodef{NMEA}{National Marine Electronics Association}
\acrodef{PPK}{Post Processed Kinematic}
\acrodef{RTK}{Real Time Kinematic}
\acrodef{INS}{Inertial Navigation System}
\acrodef{GT}{Ground Truth}
\acrodef{ICP}{Iterative Closest Point}
\acrodef{CAD}{Computer Aided Design}
\acrodef{FoMo}[FoMo]{For\^{e}t Montmorency}
\acrodef{FMCW}{Frequency Modulated Continuous Wave}
\acrodef{MPC}{Model Predictive Controller}
\acrodef{GNSS}{Global Navigation Satellite System}
\acrodef{TaR} [T\&R]{Teach and Repeat}
\acrodef{VTR}[VT\&R]{Visual Teach and Repeat}
\acrodef{RTR}[RT\&R]{Radar Teach and Repeat}
\acrodef{LTR}[LT\&R]{Lidar Teach and Repeat}

\usepackage{fancyhdr}
\fancypagestyle{withfooter}{
  
  \fancyhead[L]{}
  \fancyhead[R]{}
  \fancyfoot[C]{\footnotesize Presented at the 2025 IEEE ICRA Workshop on Field Robotics}
}
\begin{document}

\maketitle
\thispagestyle{withfooter}
\pagestyle{withfooter}

%%%%%%%%%%%%%%%%%%%%%%%%%%%%%%%%%%%%%%%%%%%%%%%%%%%%%%%%%%%%%%%%%%%%%%%%%%%%%%%%
\begin{abstract}
Teach and repeat is a rapid way to achieve autonomy in challenging terrain and off-road environments. 
A human operator pilots the vehicles to create a network of paths that are mapped and associated with odometry. 
Immediately after teaching, the system can drive autonomously within its tracks.
This precision lets operators remain confident that the robot will follow a traversable route. 
However, this operational paradigm has rarely been explored in off-road environments that change significantly through seasonal variation. 
This paper presents preliminary field trials using lidar and radar implementations of teach and repeat.
Using a subset of the data from the upcoming FoMo dataset, we attempted to repeat routes that were 4 days, 44 days, and 113 days old. 
Lidar teach and repeat demonstrated a stronger ability to localize when the ground points were removed. 
FMCW radar was often able to localize on older maps, but only with small deviations from the taught path. 
Additionally, we highlight specific cases where radar localization failed with recent maps due to the high pitch or roll of the vehicle.
We highlight lessons learned during the field deployment and highlight areas to improve to achieve reliable teach and repeat with seasonal changes in the environment. 
Please follow the dataset at \url{https://norlab-ulaval.github.io/FoMo-website} for updates and information on the data release.
\end{abstract}

%%%%%%%%%%%%%%%%%%%%%%%%%%%%%%%%%%%%%%%%%%%%%%%%%%%%%%%%%%%%%%%%%%%%%%%%%%%%%%%%
\section{INTRODUCTION}

\ac{TaR} has become a popular framework for robot navigation.
Relying on a human expert to drive the initial path, the robot then repeats the trajectory autonomously using available sensor measurements of features in the robot's surroundings.
No \ac{GNSS} data is required for large-scale operations to be completed precisely. 
\ac{TaR} has been successfully tested with various platforms and sensor modalities, including planetary rovers equipped with a lidar~\cite{McManus2012}, drones with a stereo camera~\cite{Warren2019}, or passenger cars with a \ac{FMCW} radar~\cite{Burnett2022}.
However, \ac{TaR} is still subject to difficulties related to transfers from one platform to another and adapting to environmental shifts.

\begin{figure}[t]
    \centering
    \includegraphics[width=\linewidth]{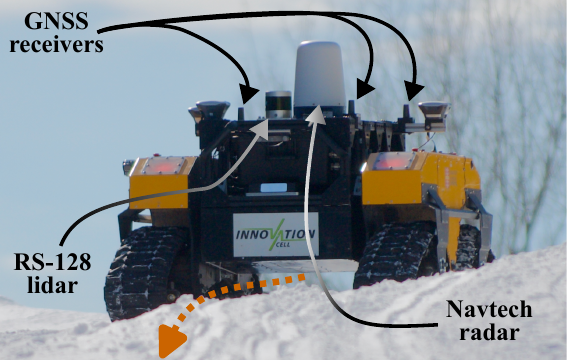}
    \caption{Our data acquisition platform in the Montmorency Forest, Quebec, where the experiments took place.
    We tested the \acl{TaR} framework using state estimation originating from lidar and radar sensors.
    }
    ~\label{fig:intro}
\end{figure}

Within the \ac{TaR} framework, a robot may encounter various environmental changes, each with differing effects and intensities.
A system operating in dense urban traffic must handle fast dynamics, such as moving vehicles and pedestrians.
Camera-based systems suffer from alternations in the scene illumination, typically occurring on a regular day-night basis~\cite{Paton2015}.
Conversely, long-term environmental changes, such as leaf fall or snow accumulation, modify the scene for most sensors~\cite{Pomerleau2023}.
For example, increasing snow cover presents new challenges in scenarios like the one depicted in \autoref{fig:intro}.
As the robot traverses the uneven snow cover, abrupt changes in the robot's roll, pitch, and attitude induce previously unseen sensor readings.
Rising snow cover can also hinder control, making maneuvers such as turning on the spot unachievable.
Even when predictable, these changes still present substantial difficulties as the sensor readings can differ significantly between the teach and repeat runs~\cite{Baril2022}.

In this work, we present a field evaluation of \ac{RTR}\footnote{\url{github.com/utiasASRL/vtr3}} and \ac{LTR}\footnote{\url{github.com/norlab-ulaval/wiln}} systems.
For our experiments, we chose a boreal forest - the largest land biome on Earth~\cite{Hayes2022}, with its dense tree canopy blocking \ac{GNSS} signal \cite{Baril2022}
and requiring a navigation solution based on the locally available sensor modalities.
Furthermore, boreal forests are subject to snow accumulation of up to several meters and environmental variations due to forestry operations or windfalls.
The main contributions of this article are:
\begin{itemize}
    \item A report on field experiments of \ac{RTR} and \ac{LTR} systems in a boreal forest, repeating trajectories that were four days, 44 days, and 113 days old.
    \item A discussion on the challenges and lessons learned from, to the best of our knowledge, the first evaluation of \ac{TaR} robustness against rising snow levels.
\end{itemize}

\section{RELATED WORK}
\acf{FMCW} radar has been gaining popularity in robotic applications, motivated by its longer range than lidar and improved consistency across weather variations~\cite{Harlow2024}. 
\acf{RTR} uses a sequence of local radar sub-maps connected topologically to localize a vehicle.
\citet{Burnett2022} introduced this concept, extracting point clouds from radar signal images and using them to perform 2D odometry and localization in a topometric map. 
They showed that the \ac{ICP} alignment of point clouds extracted from radar could be used to accurately localize a vehicle in an urban setting. 
\citet{Qiao2025} extended this idea to a complete navigation system, controlling a Clearpath Warthog \ac{UGV} in offroad environments. 
They showed that adding a gyro improved the angular velocity estimate of the vehicle compared to \ac{ICP} alone in unstructured environments. 
Their experiments included driving among trees but not the dense boreal forest considered in this experiment.
This implementation of \ac{RTR} \cite{Qiao2025} employs a \ac{MPC} that generates control commands for the vehicle to achieve high-quality path following.
Similar to radar, lidar is an active sensor, making it robust to changes in scene illumination.
Moreover, with 3D lidars becoming increasingly accessible, a navigation framework based on laser sensing inherently supports environments with changes in elevation, as well as high pitch and roll angles.
\citet{Krusi2015} demonstrated this capability of an \ac{ICP}-based \ac{LTR} in rugged unstructured outdoor terrain, as well as highly dynamic urban environments.
\citet{Burnett2022} evaluated the performance of \ac{LTR} in the context of autonomous driving, showing its robustness to moderate levels of precipitation and limited seasonal variations.

\begin{table}[b]
    \centering
    \caption{Sensor Specifications}
    \label{tab:sensors}
    \begin{tabularx}{\linewidth}{XXcr}
    \toprule
        \textbf{Sensor Type} & \textbf{Model} & \textbf{Qty.} & \textbf{Rate} \\
    \midrule
        Lidar           & RoboSense RS-128             & 1  & \SI{10}{\Hz} \\
        \ac{IMU}        & VectorNav VN-100              & 1  & \SI{200}{\Hz} \\
        \ac{GNSS}       & Emlid Reach M2                & 3  & \SI{10}{\Hz} \\
        Static \ac{GNSS}& Emlid Reach RS3               & 1  & \SI{10}{\Hz} \\
        Radar           & Navtech CIR-304H              & 1  & \SI{4}{\Hz} \\
        Wheel encoders  & Hall effect sensors           & 2  & \SI{4}{\Hz} \\
    \bottomrule
    \end{tabularx}
\end{table}

% \begin{table}[b]
%     \centering
%     \caption{Sensor Specifications}
%     \label{tab:sensors}
%     \begin{tabular}{llll}
%     \toprule
%         \textbf{Sensor} & \textbf{Model} & \textbf{Qty.} & \textbf{Rate} \\
%     \midrule
%         Lidar           & RoboSense RS-128             & 1  & \SI{10}{\Hz} \\
%         \acs{IMU}        & VectorNav VN-100              & 1  & \SI{200}{\Hz} \\
%         \ac{GNSS}       & Emlid Reach M2                & 3  & \SI{10}{\Hz} \\
%         Static \ac{GNSS}& Emlid Reach RS3               & 1  & \SI{10}{\Hz} \\
%         Radar           & Navtech CIR-304H              & 1  & \SI{4}{\Hz} \\
%         Wheel encoders  & Hall effect sensors           & 2  & \SI{4}{\Hz} \\
%     \bottomrule
%     \end{tabular}
% \end{table}

Focusing specifically on seasonal changes,~\citet{Harlow2024} discussed the robustness of \ac{FMCW} radars in various weather conditions.
\citet{Gridseth2022} employed a deep learning approach, training a neural network with camera data from summer, winter, and spring to show that predicted visual features remained effective over several months.
\citet{Rozsypalek2023} addressed seasonal challenges in \ac{VTR} using a Monte Carlo state estimation.
Although the authors include repeat scenes with snow, the snow accumulation on the ground is minimal.
In boreal forests,~\citet{Baril2022} demonstrated the ability of a \ac{LTR} framework to repeat a trajectory between seasons.
However, the authors only considered snow cover reduction, repeating multiple end-of-winter trajectories in the fall.
In contrast, this report discusses the performance of \ac{RTR} and \ac{LTR} as the snow level rises, providing insights into their capabilities under such conditions.

\section{EXPERIMENTS}

\begin{figure}[tbp]
    \centering
    \includegraphics[width=\linewidth]{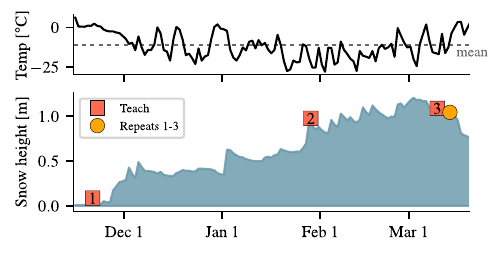}
    \caption{The average daily temperature and accumulated snow cover at the experimental site in the Montmorency Forest between the November and March sessions.
    The initial teach recording in November occurred without any snow cover, and by the time of the March experiments, the snow depth had reached \SI{1}{\meter}.
    The average temperature, denoted by the dashed line, was \SI{-11.46}{\celsius} during this period.}
    ~\label{fig:weather}
\end{figure}

All of the navigation experiments were performed during winter 2025 in the Montmorency Forest located \SI{70}{\kilo\metre} north of Quebec City, Canada, at a latitude of~\ang{47;19;15}N and a longitude of~\ang{70;9;0}W.
The snow base was over \SI{1}{\meter} with temperatures ranging from \SI{-26}{\celsius} to \SI{5}{\celsius}.
We report the average daily temperature together with the snow cover data in \autoref{fig:weather}.

\subsection{Robot Platform and Sensor Setup}

The experiments were conducted using a Clearpath Robotics Warthog UGV, as depicted in \autoref{fig:intro}.
This vehicle features four CAMSO ATV T4S tracks instead of traditional wheels, enhancing its capabilities in deep snow.
The skid-steered mobile platform is equipped with a differential suspension system.
Its advanced battery system comprises 16 Lithium-Ion battery modules, providing a total capacity of \SI{7.8}{\kilo\watt\hour}.
Additionally, a custom-built modular sensor frame is mounted on the robot's chassis.

\begin{table*}[t]
    % \centering
    \caption{
        Multi-Season Autonomous Capability of \acl{RTR}.        
    }
    \vspace*{-0.35cm}
    \parbox{18cm}{\centering
    The reported autonomy rate is computed with respect to the trajectory repeat duration, while the path lengths come from \ac{GNSS} data.
    }
     \begin{tabularx}{\textwidth}{X  X  X  X  X  X  X}
    \toprule
        \textbf{Teach} &
        \textbf{Autonomy Rate} & 
        \textbf{Autonomy Length} & 
        \textbf{Path Length}  & 
        \textbf{Snow Height}& 
        \textbf{Time to Repeat} \\ 
    \midrule
        Nov. 21, 2024 & \SI{84.1}{\percent} & \SI{218}{\meter} & \SI{577}{\meter} & \SI{0.00}{\meter} & 114 days\\
        Jan. 29, 2025 & \SI{91.8}{\percent} & \SI{558}{\meter} & \SI{570}{\meter} & \SI{0.92}{\meter} & 44 days \\
        Mar. 13, 2025 & \SI{100.0}{\percent} & * & * & \SI{1.07}{\meter} & 1 day \\
    \bottomrule
    \end{tabularx}
    \footnotesize{\\\\Note: * denotes missing \ac{GNSS} data.}
    ~\label{tab:radarRepeatInfo}
\end{table*}

The experiments in this field deployment use only a small subset of the available sensor modalities in the FoMo Dataset~\cite{Boxan2024}. 
The \acl{LTR} under evaluation~\cite{Baril2022} relies on the RoboSense RS-128 lidar, the VectorNav VN-100 \ac{IMU} and wheel encoders.
Our \acl{RTR}~\cite{Qiao2025}, on the other hand, uses the Navtech CIR-304 radar and the gyro measurements from the VectorNav VN-100 \ac{IMU}.
Precise \ac{GNSS} position is acquired with three Emlid M2 receivers mounted on the \ac{UGV}, while a single Emlid Reach RS3 serves as a static reference station.
The information from all four receivers is post-processed into 6-degree-of-freedom trajectories for use in evaluation.
Detailed sensor specifications can be found in~\autoref{tab:sensors}.

\subsection{Experimental Environment}

\begin{figure}[!b]
    \centering
    \includegraphics[width=1.0\linewidth]{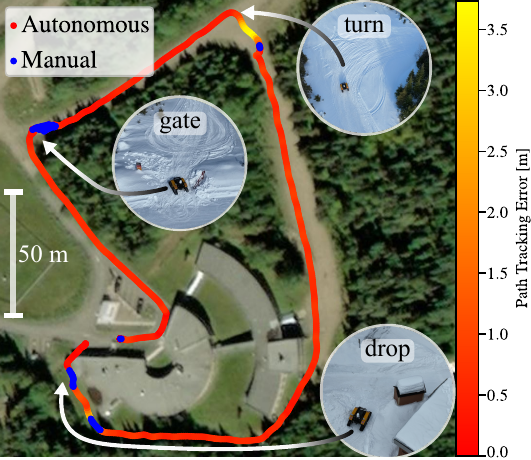}
    \caption{Path Tracking Error for a \ac{RTR} run on the \emph{blue} trajectory recorded in January and repeated in March.
    The blue sections of the trajectory correspond to the data points where the robot was controlled manually.
    The photo insets depict challenging spots on the trajectory.
    At the \emph{gate}, the robot transitions from the plowed road to a forest trail, passing through a narrow gate as it climbs a snow bank.
    At the \emph{turn}, the robot turns by about \SI{90}{\degree} in snow.
    The \emph{drop} location features a narrow path between two snow banks that were not present during the November teach.
    Satellite image from \url{https://mapy.cz}.
    }
    ~\label{fig:map_trajectory_error}
\end{figure}

For our experiments, we chose the \emph{blue} trajectory from the FoMo~dataset.
The trajectory, depicted in~\autoref{fig:map_trajectory_error}, is \SI{570}{\meter} long with altitude difference of \SI{11}{\meter}.
Robot pitch and roll angles varies in seasons and snow level.
In March, we reported values between \SI{-11.5}{\degree} and \SI{8.1}{\degree} for pitch and between \SI{-14.4}{\degree} and \SI{10.6}{\degree} for roll.
During this trajectory, the robot starts next to the area's main building, which offers sufficient features for initial localization.
The robot then proceeds along a plowed road with tall snowbanks before entering a nearby forest through a gate.
In this \emph{gate} location, the robot has to pass through a narrow spot as it climbs up a snow bank, reaching an unplowed forest trail.
Later along the \emph{blue} trajectory, the robot reaches the \emph{turn} location, where it turns on the spot before looping back around the other side of the main building.
Finally, the robot returns to the road through a narrow path between two snow banks at the \emph{drop} location.
Before our experiments, we flattened the unplowed snow trail with a snowmobile.
The snow packing helped lower the platform's power consumption and improve the robot's mobility, as our \ac{MPC} was not fine-tuned for deep snow. 

\begin{figure*}[tbp]
    \centering
    \begin{overpic}[width=0.98\textwidth]{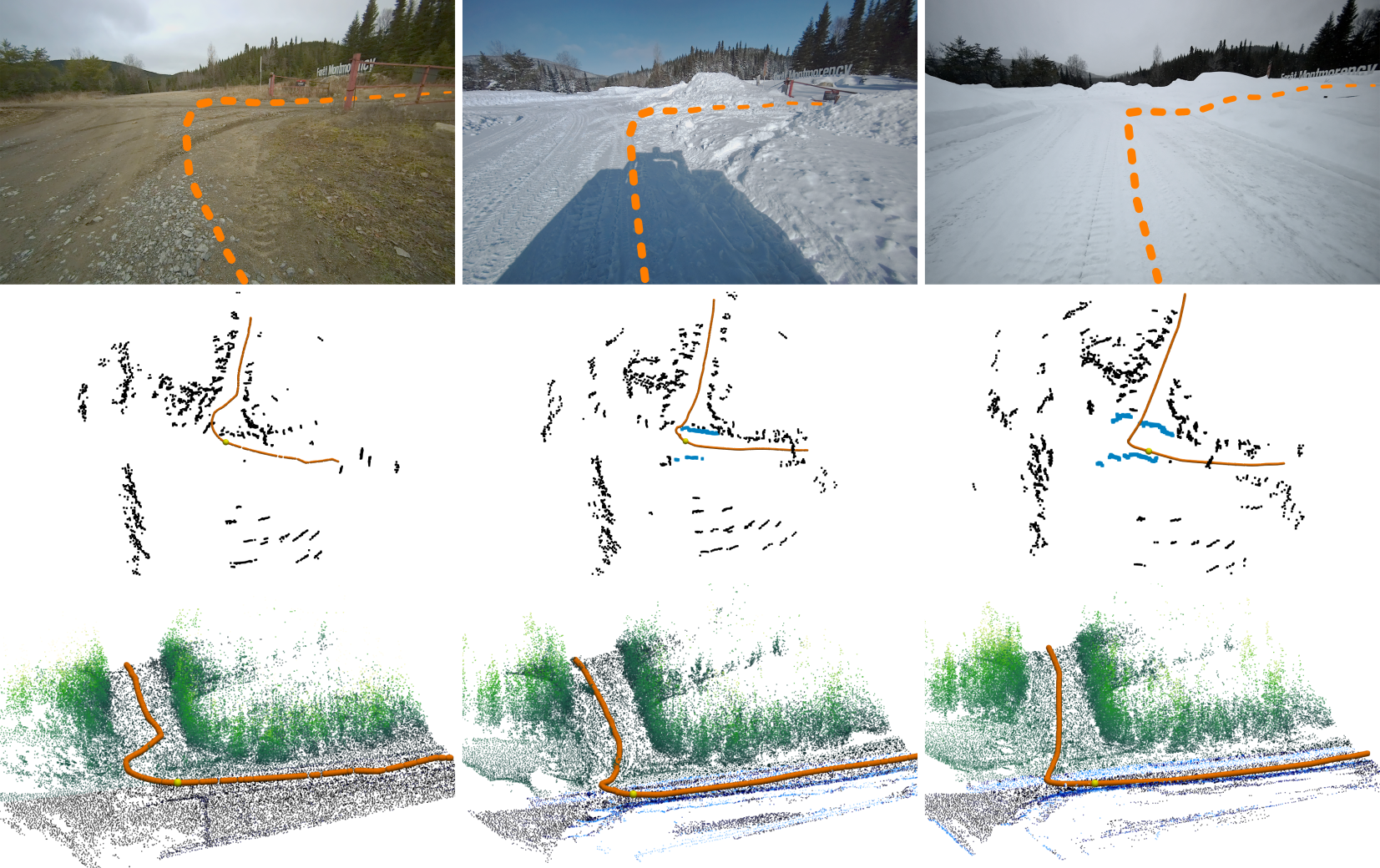}
        \put(10,61){\color{black}\text{Nov. 21, 2024}}
        \put(43,61){\color{black}\text{Jan. 29, 2025}}
        \put(77,61){\color{black}\text{Mar. 10, 2025}}
    \end{overpic}
    \caption{Views on the \emph{gate} location on the \emph{blue} path taken during three different time points (in columns).
    The first row shows images from the front-facing camera mounted on the robot.
    The second row shows radar sub-maps with the robot's trajectory in orange and snowbanks highlighted in blue.
    The last row depicts a part of the global lidar point cloud, used for lidar \ac{TaR}, again with the robot's trajectory in orange.
    Green indicates height and vegetation, while the blue color in the January and March columns highlights snow accumulation around the road.
    Yellow spheres in the orange trajectories indicate the robot's position where the corresponding image in the top row was taken. 
    }
    ~\label{fig:season-sensor-comparision}
\end{figure*}

For each of our teach runs, we selected a different data recording from the~FoMo dataset.
These recordings, executed in November, January, and March, were 113 days, 44 days, and four days old, respectively, at the time of the repeat experiments.
\autoref{fig:season-sensor-comparision} illustrates the seasonal changes at the \emph{gate} location across these data recordings.
Although no snow was present in November, the snow banks around the plowed road became more pronounced in both the radar and lidar scans as snow accumulated at the experiment site during the later winter months.

Repeats were performed for six cases: lidar and radar repeats for the three months corresponding to three individual teach routes. 
Only the radar repeat using data that was taught in the same deployment was able to achieve a \SI{100}{\percent} autonomous repeat of the \textit{blue} route. 
Unfortunately, due to an issue with data logging, the ground truth \ac{GNSS} data is missing for that repeat, leaving a gap in the table for that section. 
The robot ran out of battery, corrupting logs of the lidar experiments.
For this reason, \autoref{tab:radarRepeatInfo} contains only radar information.

\section{RESULTS AND DISCUSSION}

For the January radar repeat, the \ac{GNSS} path-tracking RMSE was \SI{0.77}{\meter}.
This is much larger than the \SI{0.15}{\meter} estimated path tracking error from the radar state estimate. 
Notably, the median of the signed error is \SI{-0.572}{\meter}.
This may suggest a bias in the GNSS that changes between repeats over long time intervals. 

%\section{DISCUSSION}

The terrain at Montmorency Forest contains more significant and abrupt orientation changes than the field testing performed in the original \acl{RTR} paper~\cite{Qiao2025}. 
As the \ac{FMCW} radar is a two-dimensional sensor, it is more difficult to match scans with a change in roll or pitch between them. 
When driving in a closed loop, the assumption with radar teach and repeat is that if the robot repeats closely in its tracks, its attitude in teach and repeat will be similar enough to allow scan matching to localize the robot.
Yet, experiments conducted with a gap of 113 days and \SI{1.04}{\meter} of accumulated snow mean that the robot cannot follow the same 3D trajectory. 
However, even in repeats performed less than one hour after teaching, the variability in scans due to path-tracking errors led to faults. 

\begin{figure*}[tbp]
    \centering
    \includegraphics[width=\textwidth]{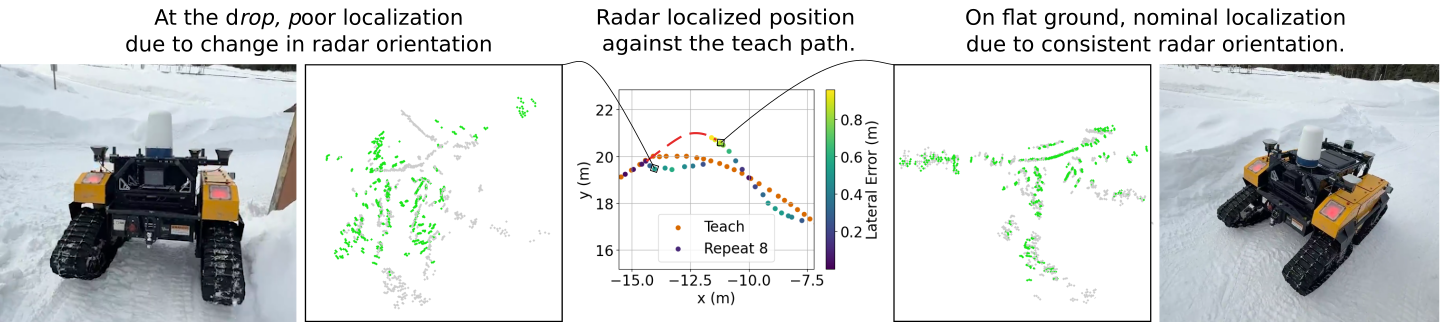}
    \caption{A comparison of the radar localization as the Warthog \ac{UGV} travels through the \emph{drop}, on the left, to the plowed path, on the right. The live radar scan at each location is colored green compared to the local submap in grey. The live scan and map are misaligned at the \emph{drop} but aligned well on the flat ground. The center panel shows that the localization estimate is false to the right of the path, causing the robot to drive off the path (shown in dashed red). Once localization recovers, the robot returns to the path. }
    ~\label{fig:dropFig}
\end{figure*}

We have identified two locations where snowbanks caused abrupt pitch changes: the \emph{gate} and the \emph{drop}, depicted in \autoref{fig:map_trajectory_error}.
These two locations were transition points between plowed roads and forest trails with compacted but unaltered snow accumulation.
To increase the difficulty further, the robot turns by about \SI{90}{\degree} at these locations as well, meaning that the control problem is also more difficult at the same time as localization struggles. 
\Cref{fig:dropFig} shows the Warthog as it enters the \textit{drop} region. 
As it drives over the hill, the view from the radar changes, and it detects the ground differently than in the teach. 
The orientation discrepancy causes the controller to drive the vehicle off the target path. 

To evaluate the multi-season capabilities of \ac{RTR}, autonomous repeats were attempted using maps from the January and November FoMo data recordings. 
When repeating the January route, the robot never lost localization while driving. 
However, manual intervention was required at the \textit{gate} because the snow drifts made the path taken in January too difficult to drive in March.
For the January repeat, the areas with the largest errors (in yellow in \autoref{fig:map_trajectory_error}) occurred right before or after manual intervention due to slippage or poor path following.
The November route was less successful. 
The November map only allowed the robot to drive until the \textit{turn}. 
A manual intervention was attempted to realign the Warthog to the path, but localization did not recover on its own, so the repeat was canceled at that point. 
The magnitude of change between November, recorded without snow, and March, with \SI{1.04}{m} of snow accumulation, leads to significant changes in the features detected by the radar. 
Snow banks are visible to the radar and change the results. 
Additionally, kinematically, the vehicle has changed from wheels to tracks. 
In the November path, there is a turn on the spot at the \textit{turn} (See \autoref{fig:map_trajectory_error}).
With treads on deep snow, the Warthog cannot turn on the spot. 
The tight coupling of the localization and motion means that when motion is difficult, the localization process struggles because it does not anticipate that the robot will deviate from the path. 
In this case, localization did not recover after a manual intervention to move the robot further along the path. 
Overall, localization is close to operating reliably for radar through this seasonal change, but additional terrain assessment will be required because the traversability can change drastically over such a large time change. 

\acl{LTR} demonstrated better performance in managing the additive changes in the environment, thanks to the 128-beam~3D data provided by the RS-128 lidar.
As the employed \ac{LTR} system~\cite{Baril2022} processes trajectories in two dimensions, the controller effectively disregards variations in the robot's attitude caused by snow accumulation.
Consequently, the primary issue observed with lidar localization occurred during the repeat of the 113-day-old November run.
During this repeat, the \ac{ICP}-based lidar localization failed in areas with plowed snow due to snow banks reaching up to three meters in height.
To address this, we applied a bounding box filter on the lower part of the scan point cloud, effectively removing all points lower than \SI{1.5}{\meter} over the lidar.
Due to the high resolution and wide vertical field of view of the employed lidar, the localization algorithm still had enough features from trees and other tall objects to localize successfully.
This approach, however, may not be efficient in environments with high snow cover but fewer tall features, such as tundra or ice sheet biomes.
Future work could explore more advanced filtering techniques, such as semantic segmentation and point cloud projection on camera images from the FoMo dataset.

The main challenges encountered with \ac{LTR} involved difficulties with controlling a skid-steered tracked vehicle on snow.
Similar to \ac{RTR}, the \emph{gate} location was particularly challenging as the trajectory followed a highly dynamic path over a snow bank.
A later section of the blue trajectory included a \SI{90}{\degree} \emph{turn}, see \autoref{fig:map_trajectory_error}.
Such sharp turns are difficult to execute on snow and should be avoided during teach runs, as smoothing and filtering the resulting trajectory might not yield optimal results, particularly when the inner part of the turn contains obstacles.
Similarly, high snow banks around plowed roads prevented the exact repetition of taught trajectories, especially in turns.
Therefore, we conclude that future \ac{TaR} systems need basic obstacle avoidance capabilities and terrain assessment to ensure multi-season operations.

Finally, we report several general lessons learned during our field experiments.
A limiting factor for mobile robot deployment, especially in winter, is battery capacity.
Our experiments indicate that a tracked platform consumes approximately twice as much power as a wheeled one.
The lack of sunlight in the winter months reduces deployment time, and experiments must be scheduled to allow sufficient charging time.
Additionally, users should consider the current snow and meteorological conditions in their planning.
Deep or wet snow significantly limits a robot's energy autonomy, and conditions can change rapidly due to sunshine melting the top layers of snow, particularly in spring.

\section{CONCLUSION AND FUTURE WORK}
This report presented the results, challenges, and lessons learned from a field test of \ac{RTR} and \ac{LTR} in a boreal forest.
Our experiments, including repeats of 113 days, 44 days, and 4 days-old trajectories, showed that high accumulation of snow still presents a challenge for successful localization.
Snow banks, in particular, showed the need to couple multiple data modalities together, as both lidar and radar localization matched non-existing features.
An important question for evaluating radar odometry in off-road environments is how to properly account for 2D information in a 3D space. 
As the robot moves, if the odometry is locally correct, the total distance would be along the path manifold, longer than its projection in 2D. 
However, because we are not aware of this 3D motion in the first place, it is not possible to perform the projection. 
This makes comparison between radar estimates and ground truth more difficult. 
We plan to address this issue explicitly in the automated evaluation of the FoMo~dataset once it is released.

\printbibliography

\end{document}